\documentclass[10pt,conference]{IEEEtran}
\IEEEoverridecommandlockouts
\usepackage{cite}
\usepackage{amsmath,amssymb,amsfonts}
\usepackage{algorithmic}
\usepackage{graphicx}
\usepackage{textcomp}
\usepackage{xcolor}

\usepackage{booktabs}
\usepackage[hidelinks]{hyperref}

\def\BibTeX{{\rm B\kern-.05em{\sc i\kern-.025em b}\kern-.08em
    T\kern-.1667em\lower.7ex\hbox{E}\kern-.125emX}}
\begin{document}

\title{Causal-INSIGHT: Probing Temporal Models to Extract Causal Structure}

\author{
\IEEEauthorblockN{
Benjamin Redden,
Hui Wang$^{*}$,
Shuyan Li
}
\IEEEauthorblockA{
School of Electronics, Electrical Engineering and Computer Science\\
Queen's University Belfast, Belfast, United Kingdom\\
\{bredden01, h.wang, shuyan.li\}@qub.ac.uk
}
}

\maketitle

\begin{center}
\footnotesize
© 2026 IEEE. Accepted for publication at IJCNN 2026.
\end{center}

\begingroup
\renewcommand{\thefootnote}{}
\footnotetext{$^{*}$Corresponding author: h.wang@qub.ac.uk}
\endgroup

\begin{abstract}
Understanding directed temporal interactions in multivariate time series is essential for interpreting complex dynamical systems and the predictive models trained on them. We present Causal-INSIGHT, a model-agnostic, post-hoc interpretation framework for extracting model-implied (predictor-dependent), directed, time-lagged influence structure from trained temporal predictors. Rather than inferring causal structure at the level of the data-generating process, Causal-INSIGHT analyzes how a fixed, pre-trained predictor responds to systematic, intervention-inspired input clamping applied at inference time.

From these responses, we construct directed temporal influence signals that reflect the dependencies the predictor relies on for prediction, and introduce Qbic, a sparsity-aware graph selection criterion that balances predictive fidelity and structural complexity without requiring ground-truth graph labels. Experiments across synthetic, simulated, and realistic benchmarks show that Causal-INSIGHT generalizes across diverse backbone architectures, maintains competitive structural accuracy, and yields significant improvements in temporal delay localization when applied to existing predictors.
\end{abstract}


\section{Introduction}

Deep learning models are widely used to model complex multivariate time series in domains such as neuroscience, healthcare, finance, and climate science. Despite strong predictive performance, these models are typically deployed as black boxes, offering limited insight into the temporal dependencies and interactions they learn. This lack of interpretability is particularly problematic in high-stakes settings, where understanding \emph{why} a model makes certain predictions is often as important as predictive accuracy.

A central aspect of temporal interpretability is understanding \emph{directed influence over time}: which variables affect others, and with what delay. In time series analysis, such dependencies are commonly formalized under Granger-style assumptions, where a variable is considered causally relevant if its past improves the prediction of another variable beyond other observed histories \cite{grangerOriginal}. In modern temporal modeling pipelines, predictive performance and causal analysis are often decoupled, motivating post-hoc methods for interpreting dependencies learned by trained temporal predictors.

In this work, we address this gap by reframing temporal causal analysis as a post-hoc model interpretability problem. Instead of inferring causal structure directly from observational data or inspecting model internals, we estimate dependencies encoded by a trained temporal predictor by analyzing its responses to controlled input clamping at inference time.

We introduce Causal-INSIGHT (\textbf{IN}tervention-inspired \textbf{S}ignal \textbf{I}nference for causal \textbf{G}raph discovery from \textbf{H}istorical \textbf{T}ime series), a model-agnostic framework that analyzes how the effects of input clamping propagate through a trained temporal predictor over time. These responses are aggregated into interpretable influence signals, from which we construct directed temporal graphs using \emph{Qbic}, a new sparsity-aware, graph selection criterion that balances predictive utility against structural complexity. Causal-INSIGHT operates purely at inference time and treats predictors as black boxes, enabling uniform application across architectures without modifying models or training procedures.

Causal-INSIGHT does not claim identifiability of the underlying data-generating causal graph without additional assumptions. Instead, it recovers the directed temporal dependency structure that a trained predictor relies on under Granger-style assumptions. 

Code is available at: \url{https://github.com/BenRedd/Causal-INSIGHT}.

\begin{figure*}[t]
  \includegraphics[width=\textwidth]{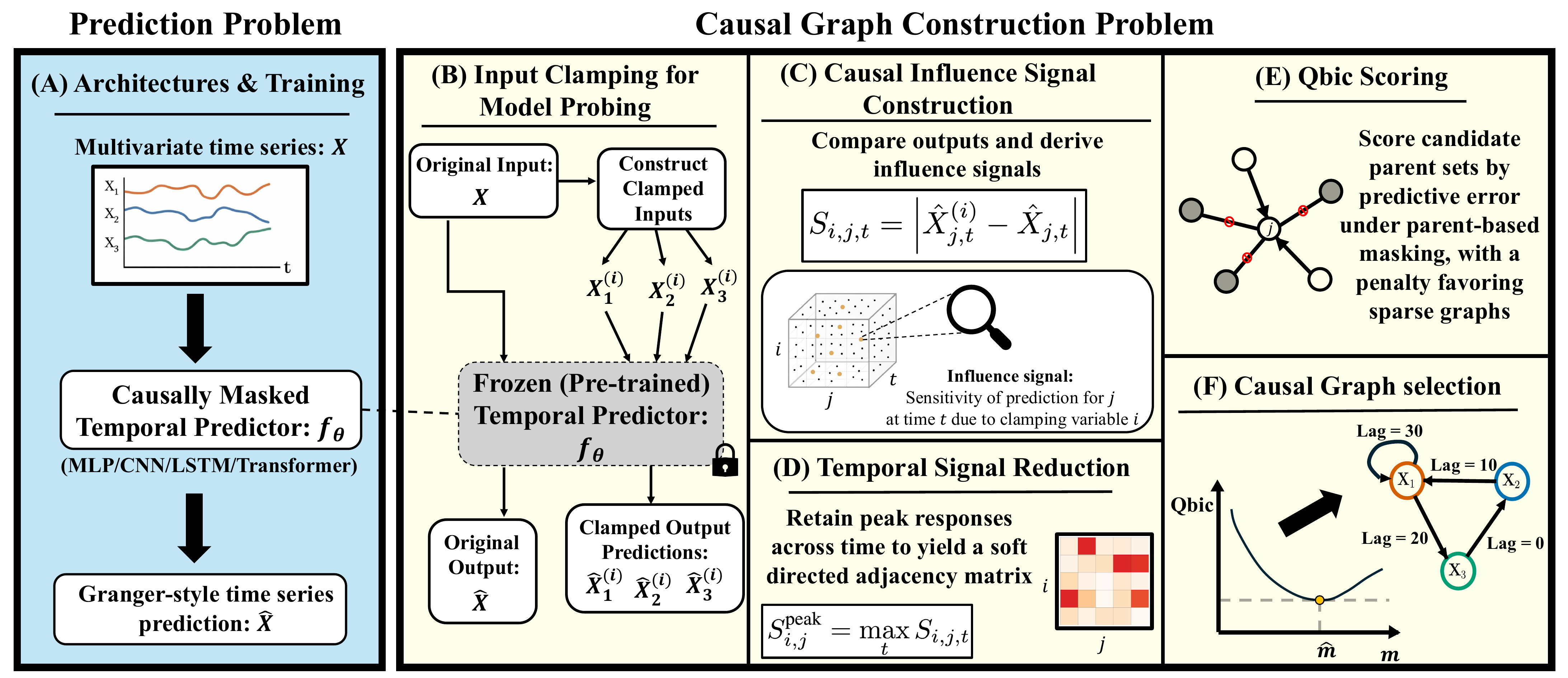}
  \caption{Overview of Causal-INSIGHT.
The method is organized into two stages: (left) the prediction problem, and (right) the causal graph construction problem. The analysis proceeds as follows:
(A) Training of a temporal predictor $f_\theta$
 under causal (no-leakage) masking constraints.
(B) Model probing via input clamping at inference time with the predictor held fixed.
(C) Construction of causal influence signals by comparing clamped and unclamped predictions.
(D) Temporal reduction of influence signals to obtain a soft structural adjacency with associated peak lags.
(E) Qbic scoring of candidate graphs using parent-based masking with a sparsity penalty.
(F) Selection of the final directed causal graph by minimizing Qbic over sparsity levels.}
\label{fig:method_overview}
\end{figure*}

\section{Related Work}

\subsection{Causal Discovery from Observational Time Series.}
Classical approaches such as Granger causality \cite{grangerOriginal} and vector autoregressive (VAR) models \cite{sims1980macroeconomics} infer directed temporal dependencies directly from multivariate time series under linearity and stationarity assumptions. Nonlinear extensions include transfer entropy \cite{schreiber2000measuring} and convergent cross mapping \cite{sugihara2012detecting}, which are explicitly designed for time-resolved dynamical systems.

Constraint-based methods such as PC \cite{spirtes1991fast} and FCI \cite{spirtes2001causation} were originally developed for cross-sectional data, but have been extended to time series settings through lagged variable representations and temporal conditional independence testing. Notably, PCMCI \cite{runge2019detecting} and its extensions PCMCI+ \cite{pmlr-v124-runge20a} and J-PCMCI+ \cite{gunther2023jpcmci} are explicitly designed for multivariate time series and scale to high-dimensional temporal systems.

Score-based approaches such as NOTEARS \cite{notears} originate in static settings, while DYNOTEARS \cite{dynotears} extends this framework to dynamic, time-lagged graphs. Functional causal models including ANMs \cite{hoyer2008nonlinear} and TiMINo \cite{timino} likewise have temporal variants that impose assumptions on functional form and noise structure to enable causal discovery from time series.

These methods aim to recover causal structure at the level of the data-generating process under assumptions such as stationarity and causal sufficiency, which are often strained in higher dimensional and noisy settings.
Moreover, these approaches operate independently of any trained predictive model, and therefore do not address which dependencies a learned temporal predictor actually relies on, nor how such dependencies are encoded in its representations.

\subsection{Deep Learning for Temporal Causal Modeling.}
Recent work extends Granger-based reasoning to nonlinear and high-dimensional settings using deep neural networks. Neural Granger Causality (with cMLP and cLSTM) \cite{tank2022neural}, eSRU \cite{khanna2020economy}, and related RNN-based methods \cite{wang2018rnn} introduce sparsity or architectural constraints during training to identify causal dependencies. Other approaches incorporate causal objectives directly into learning, including TCDF \cite{nauta2019causal}, ACD \cite{lowe2022amortized}, and graph-based models such as DVGNN \cite{dvgnn} and CUTS \cite{cuts2023}. Transformer-based architectures, most notably CausalFormer \cite{kong2025causalformer}, further leverage attention mechanisms as proxies for temporal causal structure.

While effective in specific settings, these approaches tightly couple causal inference to particular architectures, loss functions, or sparsity mechanisms introduced during training. As a result, the inferred causal structure and its interpretation are inseparable from model design choices, making these methods difficult to apply for post-hoc analysis of arbitrary predictors.

Prior work therefore addresses causal structure either at the level of the data-generating process or as an explicit component of model training. In contrast, we focus on a distinct but practically important question: how are directed temporal dependencies represented and utilized by an already trained temporal predictor, independent of its architecture or training objective?

\subsection{Post-hoc Probing and Interpretability.}
A complementary line of work studies post-hoc interpretability of trained models via attribution or perturbation-based analysis, including methods such as LIME \cite{lime} and SHAP \cite{shap}, which assess the sensitivity of model outputs to input perturbations. These methods primarily target local explanations of individual predictions and are not designed to recover global, directed temporal structure.

Causal-INSIGHT builds on the general idea of perturbation-based probing, but differs fundamentally in scope and objective. Rather than producing local attributions for individual predictions, it aggregates induced responses across variables and time to recover a global, directed, time-lagged dependency structure implied by a trained predictor. This enables systematic analysis of learned temporal representations at the graph level, without modifying model architecture or training objectives.

\section{Preliminaries}

We consider a two-stage setting, illustrated in Fig.~\ref{fig:method_overview}.
First, a \emph{prediction problem}, in which a temporal predictor is trained to estimate each variable at time \(t\) from its own past and the past and present of other variables under causal (temporal) masking constraints.
Second, a \emph{causal graph construction problem}, in which the trained predictor is analyzed post-hoc to extract directed, time-lagged influence structure.

\subsection{Prediction Problem}

Let \( \mathbf{X} = [\mathbf{X}_1, \dots, \mathbf{X}_N] \in \mathbb{R}^{N \times T} \) denote a multivariate time series with \( N \) variables and \( T \) time steps, where each row \( \mathbf{X}_i \in \mathbb{R}^T \) represents the time series for variable \( i \). The goal is to learn a parameterized function \( f_\theta \) that predicts the next value of each variable \( \mathbf{X}_i \) based on its own history and the histories of all other variables. 
We define \( \mathbf{X}_{-i, 0:t} \) as the history of all variables except \( \mathbf{X}_i \), from time step \( 0 \) to \( t \). Similarly, \( \mathbf{X}_{i, 0:t-1} \) denotes the history of the target variable up to \( t{-}1 \). The model prediction for variable \( i \) at time \( t \) is given by:
\begin{equation}
\hat{\mathbf{X}}_{i,t} = f_\theta\left( 
    \mathbf{X}_{i, 0:t-1},\, \mathbf{X}_{-i, 0:t} 
\right),
\label{eq:prediction}
\end{equation}
for \( i = 1,\dots,N \) and \( t = 1,\dots,T \). This ensures that the model has access to past observations when predicting the next value, aligning with prior work in Granger-inspired causal learning, such as TCDF. By modeling predictive dependencies, this formulation encourages the model to encode directed temporal influences that correspond to causal relations under standard predictive assumptions used in Granger-style analysis. We assume throughout this work:
(1) uniformly sampled and synchronized time series,
(2) approximate weak stationarity of the underlying processes, and
(3) causal sufficiency, i.e., the absence of unobserved confounders.
These assumptions are standard in time-series causal analysis and Granger-based discovery frameworks
\cite{spirtes2001causation, pearl2009causality, peters2017elements}.

\subsection{Causal Graph Construction Problem}

Time-series causal discovery seeks to identify how variables influence one another over time, typically represented as a directed temporal graph. We denote this structure as \( G = (V, E) \), where each node \( \mathbf{X}_i \in V \) corresponds to a time series variable, and each directed edge \( (\mathbf{X}_i \rightarrow \mathbf{X}_j) \in E \) encodes a potential causal influence from variable \( i \) to variable \( j \). The graph may contain \( m = |E| \) such edges. A directed edge with lag \( \ell \), written as \( \mathbf{X}_{i, t - \ell} \rightarrow \mathbf{X}_{j, t} \), indicates that changes in \( \mathbf{X}_i \) at time \( t{-}\ell \) influence \( \mathbf{X}_j \) at time \( t \). When \( \ell = 0 \), the influence is instantaneous (i.e., contemporaneous within the sampling interval); when \( \ell > 0 \), it reflects a delayed (lagged) effect.

Unlike static causal models, which typically assume acyclicity, temporal systems often contain cycles and feedback loops. Our framework accommodates this by allowing cyclic dependencies and self-loops to capture autoregressive behavior.

\section{Methodology}

Causal-INSIGHT extracts structure from a trained multivariate temporal predictor by probing its input-output behavior under controlled input clamping. The resulting graph summarizes the dependencies the predictor relies on for prediction and is interpreted throughout as a model-implied influence structure. Under standard Granger-style assumptions, this structure corresponds to causal influence in the data-generating process. Otherwise, it provides an interpretable summary of the model’s learned temporal dependencies.

We first train a temporal predictor \( f_\theta \) under causal masking constraints, ensuring that predictions at time \( t \) depend only on past and contemporaneous inputs. The trained model is then treated as a black box and probed via input clamping at inference time. By measuring how the resulting model responses propagate through the outputs over time, we construct directed influence signals across variable pairs and temporal lags, which are subsequently filtered and selected to form a sparse temporal graph (Fig.~\ref{fig:method_overview}).

\subsection{Architectures \& Training}
\label{sec:architectures}

All backbone predictors share a common causal masking front-end that enforces no temporal leakage.
Specifically, when predicting variable \( j \) at time \( t \), the model may access all variables at lags \( 1,\dots,K{-}1 \), may access other variables at lag~0, but is prohibited from accessing its own value at lag~0, preventing self-leakage.
This masking layer is purely structural and identical across all architectures.

Building on this shared masked representation, we vary only the prediction head, allowing us to study the effect of different temporal inductive biases while keeping causal constraints fixed.
We consider four representative backbone families:

\paragraph{MLP}
A multilayer perceptron, serving as a minimal baseline.

\paragraph{CNN}
A temporal convolutional predictor capturing local temporal patterns via convolutional filters.

\paragraph{LSTM}
A recurrent predictor modeling longer-range dependencies through gated recurrence.

\paragraph{CausalFormer}
A transformer-based temporal model proposed for causal discovery, treated here purely as a black-box predictor.

All models are trained to predict the value of each variable at time \( t \) using temporally masked inputs from the multivariate history, optimized using mean squared error loss.
Training configurations are fixed within each backbone family.
After training, models are frozen and used unchanged during inference for probing with Causal-INSIGHT. Consequently, it introduces no additional learnable parameters and does not increase overfitting risk beyond that of the underlying predictor.

\subsection{Input Clamping for Model Probing}
\label{sec:input_clamping}

We analyze trained predictors using intervention-inspired input clamping, a \emph{model-probing} operation applied at inference time to measure how localized input changes propagate through a temporal predictor. The resulting graphs represent \emph{model-implied} Granger influence rather than identifiable structural causal models.

Given an observed multivariate time series \( \mathbf{X} \in \mathbb{R}^{N \times T} \), we generate a family of alternate inputs by systematically clamping a single variable at a reference time index.
For each variable \( i \), we construct
\begin{equation}
\mathbf{X}^{(i)}_{k,t} =
\begin{cases}
x^*, & \text{if } k=i \text{ and } t=t_0, \\
\mathbf{X}_{k,t}, & \text{otherwise},
\end{cases}
\end{equation}
where \( t_0 \) is fixed to \( t_0 = 0 \) in all experiments.
All other variables and time indices remain unchanged.

Under approximate stationarity, clamping any time index yields equivalent lag-aligned responses up to boundary effects. Delays are therefore estimated by measuring response timing relative to the clamping index \( t_0 \). We choose \( x^* \) to represent a strong but plausible deviation within the empirical range of each variable. For most datasets, we set \( x^* = \max(\mathbf{X}_i) \). For domains where zero has semantic meaning (e.g., fMRI), we use \( x^* = 0 \). When inputs are normalized to \( [0,1] \), this yields interpretable extremal clamping values. Because Causal-INSIGHT clamps one variable at a time, it primarily captures marginal influences and may miss higher-order interactions that manifest jointly across variables. Extending to multi-variable interventions introduces combinatorial complexity.

\subsection{Causal Influence Signal Construction}

We first compute a baseline prediction by applying the trained predictor
to the unclamped input:
\begin{equation}
\hat{\mathbf{X}} = f_\theta(\mathbf{X}),
\end{equation}
where \( \hat{\mathbf{X}} \in \mathbb{R}^{N \times T} \) contains predictions
for all variables over time. Next, for each variable \( i \), we apply intervention-inspired input clamping
to obtain a clamped input \( \mathbf{X}^{(i)} \) and compute the corresponding
predictions:
\begin{equation}
\hat{\mathbf{X}}^{(i)} = f_\theta(\mathbf{X}^{(i)}).
\end{equation}

We define the \emph{influence signal tensor}
\( \mathbf{S} \in \mathbb{R}^{N \times N \times T} \) as the absolute deviation
between clamped and baseline predictions:
\begin{equation}
S_{i,j,t} = \left| \hat{X}^{(i)}_{j,t} - \hat{X}_{j,t} \right|.
\end{equation}
We use absolute deviation to capture the magnitude of the model’s response independent of sign, which is appropriate when aggregating influence strength across heterogeneous predictors.
The entry \( S_{i,j,t} \) quantifies the sensitivity of the model’s prediction
for variable \( j \) at time \( t \) to clamping applied to variable \( i \).
Larger values indicate stronger model-implied directed influence from \( i \) to \( j \)
at delay \( t \).

\subsection{Temporal Signal Reduction}

To summarize the temporal influence profile for each variable pair, we compute the peak deviation across time:
\begin{equation}
S^{\text{peak}}_{i,j} = \max_t S_{i,j,t}.
\end{equation}

This yields a matrix \( \mathbf{S}^{\text{peak}} \in \mathbb{R}^{N \times N} \) capturing the strongest observed influence from variable \( i \) to variable \( j \).

For evaluation, we report a single dominant lag per variable pair, though the full temporal profile \( S_{i,j,:} \) may be retained to represent distributed or multi-lag effects. Candidate edges are obtained by ranking entries of \( \mathbf{S}^{\text{peak}} \) and selecting the top \( m \) values.
Each candidate corresponds to a directed influence at a specific time lag.
For each unordered pair \( \{i,j\} \), we retain at most one directed edge, selecting the direction with the larger peak influence magnitude. This avoids ambiguous bidirectionality and enforces a single dominant interaction per pair. Self-loops are permitted.

\subsection{Qbic Scoring}

Selecting an appropriate graph sparsity level requires balancing predictive utility against structural simplicity. We introduce Qbic, a heuristic graph-label-free score that evaluates candidate graphs with respect to a fixed, trained temporal predictor. The name Qbic (Quality-BIC) reflects its analogy to Bayesian information criteria \cite{bic}, combining data fit and complexity penalties (although we do not imply likelihood-based interpretation). Qbic scores a candidate graph by measuring how well its inferred parent sets preserve the predictor’s behavior on observed time series data. For a candidate graph \( \mathcal{G}^{(m)} \) with \( m \) edges, Qbic is defined as
\begin{equation}
\mathrm{Qbic}\!\left(\mathcal{G}^{(m)}\right)
=
\sum_{j=1}^{N}
\left[
n \log\!\left(\mathrm{MSE}_j\right)
+
\lambda \, k_j \log(n)
\right],
\label{eq:qbic}
\end{equation}
where \( n \) is the number of valid prediction time points,
\( \mathrm{MSE}_j \) is the mean squared error for variable \( j \) when predicted using only its inferred parents,
\( k_j \) is the in-degree of node \( j \),
and \( \lambda > 0 \) controls the sparsity penalty.

To compute \( \mathrm{MSE}_j \), we evaluate the trained predictor using only the parent variables \( \mathrm{Pa}(j) \), implemented via input masking at inference time:
\begin{equation}
\hat{X}_j
=
f_\theta\!\left(
\mathrm{Mask}\!\left(\mathbf{X}; \mathrm{Pa}(j)\right)
\right).
\label{eq:parent_mask}
\end{equation}
If the inferred parent set captures the dominant dependencies encoded by the predictor, masking non-parent variables preserves predictive accuracy, while omitting relevant parents increases error.

Because all candidate graphs are evaluated under the same masking protocol and fixed predictor, Qbic values are directly comparable across sparsity levels and provide a practical criterion for sparsity selection.

\subsection{Causal Graph Selection}

We evaluate Qbic over candidate edge counts and select the sparsity level
\begin{equation}
\hat{m} = \arg\min_m \mathrm{Qbic}\!\left(\mathcal{G}^{(m)}\right),
\end{equation}
using a conservative selection rule that favors simpler graphs when multiple sparsity levels yield comparable scores.

For each retained directed edge \( (i \rightarrow j) \), the associated temporal delay is estimated as
\begin{equation}
\ell_{i,j} = \arg\max_t S_{i,j,t},
\end{equation}
where \( S_{i,j,t} \) denotes the intervention-induced influence signal.
The resulting graph represents directed, time-lagged influence structure implied by the trained predictor under the stated assumptions.

\section{Experiments}

Although Causal-INSIGHT is framed as a post-hoc interpretability method, we evaluate the extracted influence graphs using standard causal discovery metrics.
This assesses whether model-implied temporal influence aligns with known causal structure when predictors are trained under temporal masking constraints.

\subsection{Datasets}

We evaluate Causal-INSIGHT on synthetic, simulated, and realistic benchmarks adopted from CausalFormer to ensure fair comparison.
These datasets span a wide range of dimensionalities, sequence lengths, and complexity.

\paragraph{Synthetic}
We use four canonical causal structures (fork, v-structure, mediator, diamond) with additive noise \cite{diamond,fork,mediator,v}, each instantiated with 10 simulations (40 datasets total).

\paragraph{Lorenz-96}
We evaluate 10 simulations of the chaotic Lorenz-96 system \cite{lorenz} with $N=10$ variables and sequence lengths from 50 to 5000.

\paragraph{fMRI}
We use 28 simulated fMRI datasets \cite{fMRI} with between 5-50 variables, representing noisy, high-dimensional settings with indirect measurements and temporal smoothing.

\subsection{Baselines}

We compare against representative deep learning methods for temporal causal discovery, including \textbf{cMLP}, \textbf{cLSTM}, \textbf{TCDF}, \textbf{CUTS}, and \textbf{CausalFormer}, using official implementations and author-recommended configuration settings.
These methods jointly learn predictors and causal structure during training and are the closest comparators in terms of output structure. We additionally include \textbf{PCMCI+} as a classical reference on the fMRI datasets to contextualize dataset difficulty in a high-dimensional, noisy regime where observational causal discovery assumptions are strained.

\subsection{Evaluation Metrics}

We evaluate structural accuracy using precision, recall, and F1 over directed edges.
Temporal accuracy is measured by Precision of Delay (PoD), defined as the fraction of correctly identified lags among true positive edges.

We also report Structural Hamming Distance (SHD), True Positive Rate (TPR), and False Discovery Rate (FDR).
SHD is computed on directed adjacency matrices (not Markov equivalence classes), where reversed edges count as two operations.
\cite{peters2017elements,runge2019detecting}.

\subsection{Implementation Details}

All backbones are trained with fixed configurations across datasets to assess generality rather than maximize predictive performance.
Training uses mean squared error with early stopping (20 epochs for Synthetic/Lorenz, 10 for fMRI).

Based on preliminary sweeps indicating qualitative stability across a broad range of $\lambda$ values, we fix $\lambda = 0.4$ across all experiments to avoid dataset-specific tuning. Graph sparsity is selected by identifying the first stable minimum of the Qbic trajectory, with early stopping for efficiency.

Experiments were run on Ubuntu 22.04 with an NVIDIA H100 SXM GPU (80\,GB VRAM), using PyTorch~2.1.0 and CUDA~11.8. All runs used fixed random seeds.

\section{Results \& Discussion}

\begin{table*}[t]
\centering
\setlength{\tabcolsep}{4pt}
\renewcommand{\arraystretch}{1.0}
\caption{Average structural F1 (mean $\pm$ standard deviation) across dataset categories.
Results are averaged across sub-datasets within each category.
The rightmost block reports Causal-INSIGHT with different backbone predictors.}
\label{tab:f1_scores_all}
\begin{tabular}{lccccc|cccc}
\hline
\textbf{Dataset} 
& \textbf{cMLP} 
& \textbf{cLSTM} 
& \textbf{TCDF} 
& \textbf{CUTS} 
& \textbf{CausalFormer} 
& \multicolumn{4}{c}{\textbf{Causal-INSIGHT (Ours)}} \\
& & & & & 
& \textbf{MLP} & \textbf{CNN} & \textbf{LSTM} & \textbf{CausalFormer} \\
\hline
fMRI     
& $0.55 \pm 0.15$ 
& $0.54 \pm 0.14$ 
& $0.59 \pm 0.11$ 
& $0.46 \pm 0.28$ 
& $0.66 \pm 0.09$ 
& $0.53 \pm 0.14$ 
& $0.51 \pm 0.14$ 
& $0.58 \pm 0.13$ 
& $\mathbf{0.67 \pm 0.07}$ \\

Lorenz96 
& $0.67 \pm 0.05$ 
& $0.52 \pm 0.04$ 
& $0.50 \pm 0.07$ 
& $0.54 \pm 0.03$ 
& $0.71 \pm 0.06$ 
& $0.48 \pm 0.05$ 
& $0.50 \pm 0.07$ 
& $0.55 \pm 0.05$ 
& $\mathbf{0.73 \pm 0.07}$ \\

Diamond 
& $0.56 \pm 0.08$ 
& $0.52 \pm 0.07$ 
& $0.71 \pm 0.08$ 
& $0.48 \pm 0.13$ 
& $0.67 \pm 0.07$ 
& $0.76 \pm 0.08$ 
& $\mathbf{0.84 \pm 0.10}$  
& $0.74 \pm 0.11$  
& $0.75 \pm 0.12$ \\

Mediator 
& $0.61 \pm 0.07$ 
& $0.54 \pm 0.09$ 
& $0.56 \pm 0.22$ 
& $0.59 \pm 0.13$ 
& $0.71 \pm 0.06$ 
& $0.80 \pm 0.11$ 
& $\mathbf{0.86 \pm 0.14}$ 
& $0.84 \pm 0.13$ 
& $0.83 \pm 0.13$ \\

V        
& $0.69 \pm 0.10$ 
& $0.65 \pm 0.07$ 
& $0.67 \pm 0.11$ 
& $0.56 \pm 0.12$ 
& $0.77 \pm 0.05$ 
& $0.71 \pm 0.13$ 
& $0.74 \pm 0.18$ 
& $\mathbf{0.80 \pm 0.17}$ 
& $0.77 \pm 0.10$ \\

Fork     
& $0.64 \pm 0.15$ 
& $0.64 \pm 0.09$ 
& $0.58 \pm 0.25$ 
& $0.67 \pm 0.14$ 
& $0.79 \pm 0.11$ 
& $0.80 \pm 0.13$ 
& $\mathbf{0.85 \pm 0.08}$ 
& $0.81 \pm 0.14$  
& $0.75 \pm 0.14$ \\
\hline
\end{tabular}
\end{table*}

\begin{table*}[t]
\centering
\setlength{\tabcolsep}{4pt}
\renewcommand{\arraystretch}{1.0}
\caption{Average delay identification performance measured by Precision of Delay (PoD; mean $\pm$ standard deviation).
Only methods supporting delay estimation are reported. Ground-truth delays are unavailable for fMRI.}
\label{tab:pod_scores_all}
\begin{tabular}{lccc|cccc}
\hline
\textbf{Dataset} 
& \textbf{cMLP} 
& \textbf{TCDF} 
& \textbf{CausalFormer} 
& \multicolumn{4}{c}{\textbf{Causal-INSIGHT (Ours)}} \\
& & & 
& \textbf{MLP} 
& \textbf{CNN} 
& \textbf{LSTM} 
& \textbf{CausalFormer} \\
\hline
Lorenz96  
& $\mathbf{1.00 \pm 0.00}$ 
& $0.75 \pm 0.11$ 
& $0.40 \pm 0.16$
& $0.92 \pm 0.06$ 
& $0.88 \pm 0.07$ 
& $0.98 \pm 0.03$ 
& $0.99 \pm 0.01$ \\

Diamond   
& $0.92 \pm 0.10$ 
& $\mathbf{0.93 \pm 0.11}$ 
& $0.66 \pm 0.30$
& $0.59 \pm 0.20$ 
& $0.60 \pm 0.19$ 
& $0.88 \pm 0.16$ 
& $0.86 \pm 0.30$ \\

Mediator  
& $0.96 \pm 0.08$ 
& $\mathbf{1.00 \pm 0.00}$ 
& $0.47 \pm 0.44$ 
& $0.66 \pm 0.29$ 
& $0.79 \pm 0.26$ 
& $0.77 \pm 0.18$ 
& $0.60 \pm 0.41$ \\

V         
& $0.93 \pm 0.11$ 
& $\mathbf{1.00 \pm 0.00}$ 
& $0.59 \pm 0.39$ 
& $0.46 \pm 0.29$ 
& $0.88 \pm 0.15$ 
& $0.72 \pm 0.29$ 
& $0.85 \pm 0.25$ \\

Fork      
& $0.96 \pm 0.10$ 
& $\mathbf{1.00 \pm 0.00}$ 
& $0.49 \pm 0.34$ 
& $0.41 \pm 0.27$ 
& $0.79 \pm 0.16$ 
& $0.89 \pm 0.14$ 
& $0.85 \pm 0.16$ \\
\hline
\end{tabular}
\end{table*}

\subsection{Overall Quantitative Trends}

Tables~\ref{tab:f1_scores_all} and~\ref{tab:pod_scores_all} summarize structural recovery (F1) and temporal localization (PoD) across all dataset categories and backbone architectures.

Across benchmarks, Causal-INSIGHT yields structurally meaningful graphs for all tested backbones, with performance competitive with or exceeding established deep learning baselines. For every dataset category, at least one Causal-INSIGHT variant attains the highest average structural F1 among all methods, indicating robust performance across predictor capacities, from MLPs to transformer-based models. Simpler backbones suffice for low-complexity systems, while higher-capacity architectures perform better on more challenging settings such as Lorenz-96 and fMRI.

Delay identification exhibits clear backbone-dependent trends. While PoD varies across architectures and datasets, Causal-INSIGHT achieves competitive temporal localization while maintaining higher structural F1 than competing methods. High PoD scores from TCDF and cMLP are often accompanied by lower structural accuracy, indicating a trade-off between delay precision and edge recovery. In contrast, Causal-INSIGHT provides more balanced performance, refining temporal attribution without aggressive sparsification.

\subsection{Replacing CausalFormer's Interpretation Mechanism}

Causal-INSIGHT can augment or replace post-hoc interpretation mechanisms of existing causal predictors. We focus on the CausalFormer backbone, where delay localization has been identified as a limitation.

Across datasets, applying Causal-INSIGHT to CausalFormer preserves or improves average structural F1 in all but one case (Fork), where performance remains comparable (Table~\ref{tab:f1_scores_all}). Improvements are modest and not statistically significant (median $\Delta$F1 = 0.014, $p = 0.124$), but consistently non-negative, indicating that probing does not degrade edge recovery.

In contrast, temporal localization improves substantially. Across 50 datasets with delay annotations, Causal-INSIGHT achieves a median PoD improvement of $0.30$ over the original interpretation, which is statistically significant (paired Wilcoxon test, $p = 1.9 \times 10^{-6}$). These results highlight the primary strength of the framework: refining when causal influence occurs, even when structural accuracy is already strong.
\subsection{Fine-Grained Structural Analysis on fMRI Data}

While aggregate structural F1 differences are modest, fine-grained structural metrics provide insight into \emph{how} performance differs beyond F1 alone. We therefore examine SHD, TPR, and FDR on the fMRI benchmarks (Table~\ref{tab:fmri_metrics}), which reveal complementary aspects of structural error behavior.

Compared to the top competing deep learning baselines, Causal-INSIGHT applied to the CausalFormer backbone achieves lower Structural Hamming Distance (SHD), higher True Positive Rate (TPR), and lower False Discovery Rate (FDR). This indicates that the observed improvements arise from a favorable trade-off: recovering a larger fraction of true edges while simultaneously suppressing spurious connections, rather than from trivial sparsification. Consistent with this, the inferred graphs retain comparable edge counts to the underlying predictors, suggesting more accurate parent selection rather than aggressive pruning.

We additionally report PCMCI+ as a classical observational reference to contextualize performance in this challenging setting. While PCMCI+ performs strongly on low-dimensional synthetic benchmarks where its assumptions are well satisfied, its performance degrades on fMRI data, exhibiting higher SHD and FDR than learning-based approaches. This behavior is consistent with the characteristics of fMRI time series, which involve indirect measurements, temporal smoothing, and substantial noise. In such regimes, trained temporal predictors can capture complex nonlinear dependencies that are difficult to recover using purely observational conditional independence tests.

Overall, these results clarify the structural error patterns underlying the aggregate trends reported earlier and highlight the value of post-hoc analysis when classical assumptions are strained.

\begin{table}[!tb]
\caption{Structural metrics averaged over 28 fMRI datasets. Causal-INSIGHT (CF) denotes our framework applied to the CausalFormer backbone. SHD is normalized by the number of true edges.}
\centering
\begin{tabular}{lccc}
\toprule
Method & SHD $\downarrow$ & TPR $\uparrow$ & FDR $\downarrow$ \\
\midrule
TCDF & 0.82 $\pm$ 0.27 & 0.57 $\pm$ 0.12 & 0.41 $\pm$ 0.17 \\
PCMCI+ & 0.73 $\pm$ 0.26 & 0.61 $\pm$ 0.16 & 0.31 $\pm$ 0.18 \\
CausalFormer & 0.69 $\pm$ 0.25 & 0.57 $\pm$ 0.16 & 0.27 $\pm$ 0.20 \\
Causal-INSIGHT (CF) & $\mathbf{0.67 \pm 0.23}$ & $\mathbf{0.63 \pm 0.16}$ & $\mathbf{0.23 \pm 0.21}$ \\
\bottomrule
\end{tabular}
\label{tab:fmri_metrics}
\end{table}

\subsection{Qualitative Behavior}

\begin{figure}[tbp]
  \centering
  \includegraphics[width=\columnwidth]{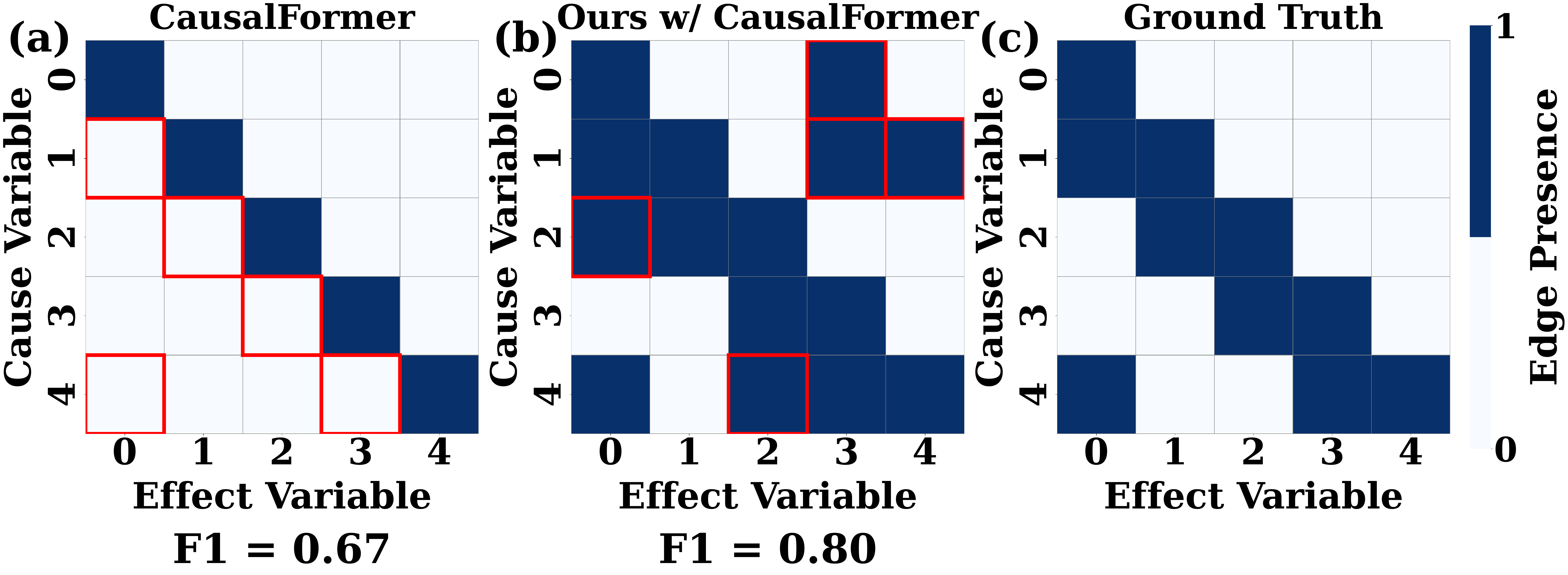}
 \caption{
Qualitative comparison of predicted adjacency matrices on a representative fMRI dataset (Sim5).
CausalFormer yields a highly conservative graph dominated by self-dependencies (diagonal),
whereas Causal-INSIGHT recovers additional directed interactions while preserving these self-loops,
resulting in higher structural accuracy (F1 $=0.80$).
Misclassifications are outlined in red.
}
  \label{fig:qualitative_adj}
\end{figure}

Fig.~\ref{fig:qualitative_adj} provides a qualitative comparison on a representative low-dimensional fMRI instance. The original CausalFormer interpretation recovers only the most dominant self-dependencies, yielding a conservative graph. In contrast, Causal-INSIGHT identifies additional directed interactions corresponding to true edges in the ground-truth graph.

While the probing-based graph is slightly denser and includes a small number of spurious edges, it recovers a substantially larger fraction of true connections, resulting in higher overall structural accuracy. This qualitative behavior mirrors the quantitative improvements observed in TPR.

\subsection{Runtime Analysis}

We measure wall-clock runtime of the Causal-INSIGHT interpretation procedure as a function of the number of variables
$N \in \{5,10,\dots,50\}$, using a fixed sequence length and a fixed backbone predictor.
For each $N$, runtime is averaged over three independently generated synthetic time series datasets constructed solely to evaluate runtime scaling behavior.

\begin{figure}[!tb]
  \centering
  \includegraphics[width=\columnwidth]{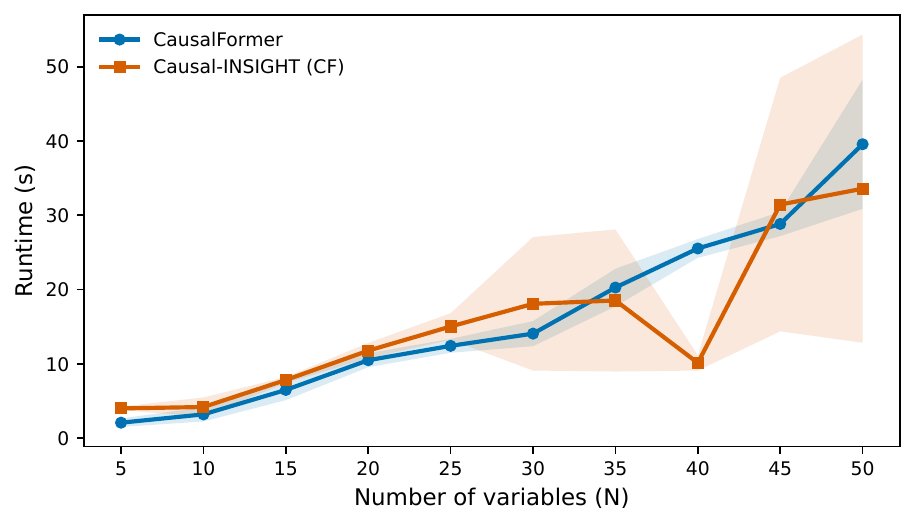}
 \caption{Runtime scaling with the number of variables.
Mean wall-clock runtime over three runs for CausalFormer and Causal-INSIGHT (CF) with fixed sequence length \(T\).
Shaded regions denote ±1 standard deviation across three runs.
Non-monotonic behavior reflects early stopping in Qbic graph selection.}
  \label{fig:runtime_scaling}
\end{figure}

As shown in Fig.~\ref{fig:runtime_scaling}, Causal-INSIGHT exhibits scaling comparable to the underlying CausalFormer predictor, with a modest constant-factor overhead.
Runtime generally increases with $N$, reflecting the cost of additional probing and graph evaluation.
The sharp drop observed around $N=40$ is due to early stopping during Qbic-based sparsity selection, which terminates graph evaluation once a stable minimum is detected.
Overall, these results indicate that the proposed post-hoc analysis preserves the empirical scaling behavior of the backbone model while incurring manageable additional cost.

For a fixed trained predictor, Causal-INSIGHT requires \(O(N)\) forward passes to construct influence signals, corresponding to one intervention per variable.
Graph selection via Qbic incurs an additional \(O(M)\) masked forward evaluations, where \(M\) denotes the number of candidate graphs evaluated before early stopping.
The overall computational cost therefore scales linearly in the number of forward passes of the backbone predictor.
Since each forward pass has the same asymptotic cost as standard inference for the chosen architecture and sequence length, the total runtime remains practical for moderate graph sizes.

\section{Ablation Studies}

\subsection{Effectiveness of Qbic for Graph Selection}

We evaluate whether Qbic provides a reliable unsupervised heuristic for selecting graph sparsity without access to ground-truth causal graphs.
Across the 28 fMRI datasets, Qbic exhibits a strong negative correlation with structural F1 score
(Pearson: $-0.77 \pm 0.36$, Spearman: $-0.53 \pm 0.43$),
indicating that lower Qbic values consistently correspond to higher-quality graph structures under a fixed predictor. On average, the graph selected by minimizing Qbic attains $91\% \pm 7\%$ of the maximum achievable F1 across candidate sparsity levels.

\begin{figure}[!tb]
  \centering
  \includegraphics[width=\columnwidth]{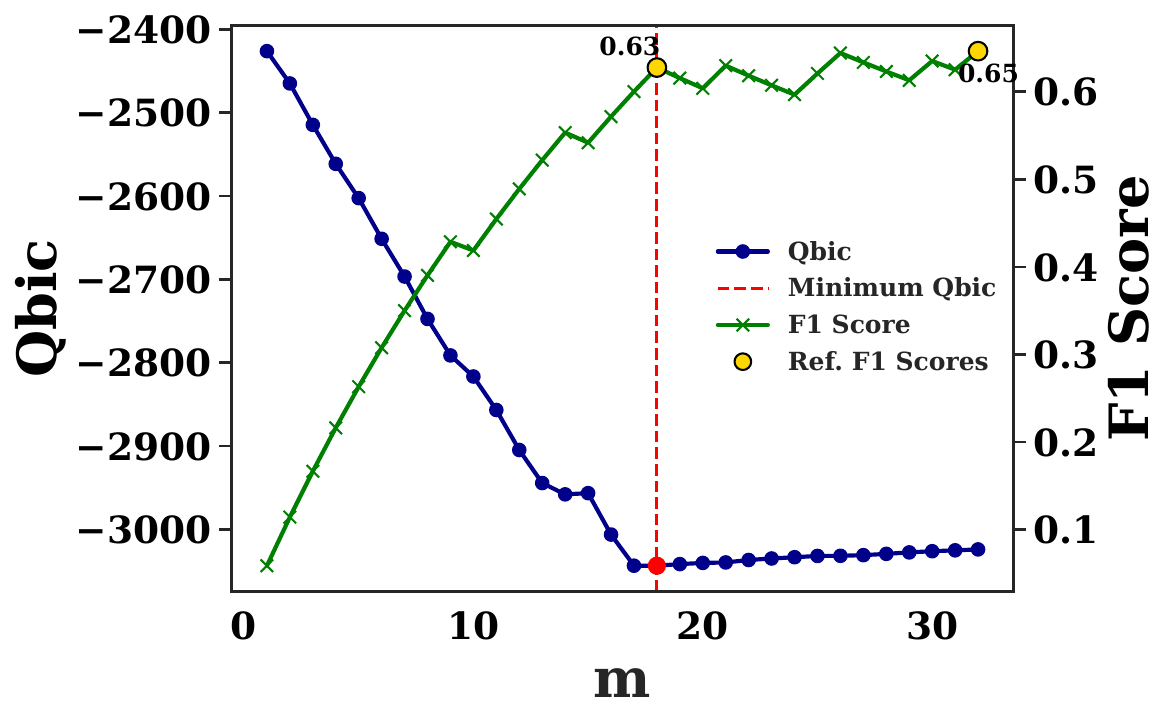}
  \caption{Qbic and structural F1 trajectories as a function of graph sparsity for a representative fMRI dataset (Sim3).
The Qbic minimum occurs at an edge count \( \hat{m} \) that yields near-maximal F1, demonstrating that Qbic reliably selects high-quality graph structures without access to ground-truth labels.}
  \label{fig:Qbic}
\end{figure}

Fig.~\ref{fig:Qbic} shows representative Qbic and F1 trajectories for an fMRI dataset, where the Qbic minimum occurs near the sparsity level that maximizes F1.

\subsection{Importance of Causal Signal Quality}

To isolate the contribution of the causal signal tensor \( \mathbf{S} \), we compare against a randomized control in which entries of \( \mathbf{S} \) are randomly permuted, preserving the marginal value distribution while destroying temporal and structural coherence. Graph construction and Qbic-based selection are then applied identically. 

\begin{figure}[!tb]
  \centering
  \includegraphics[width=\columnwidth]{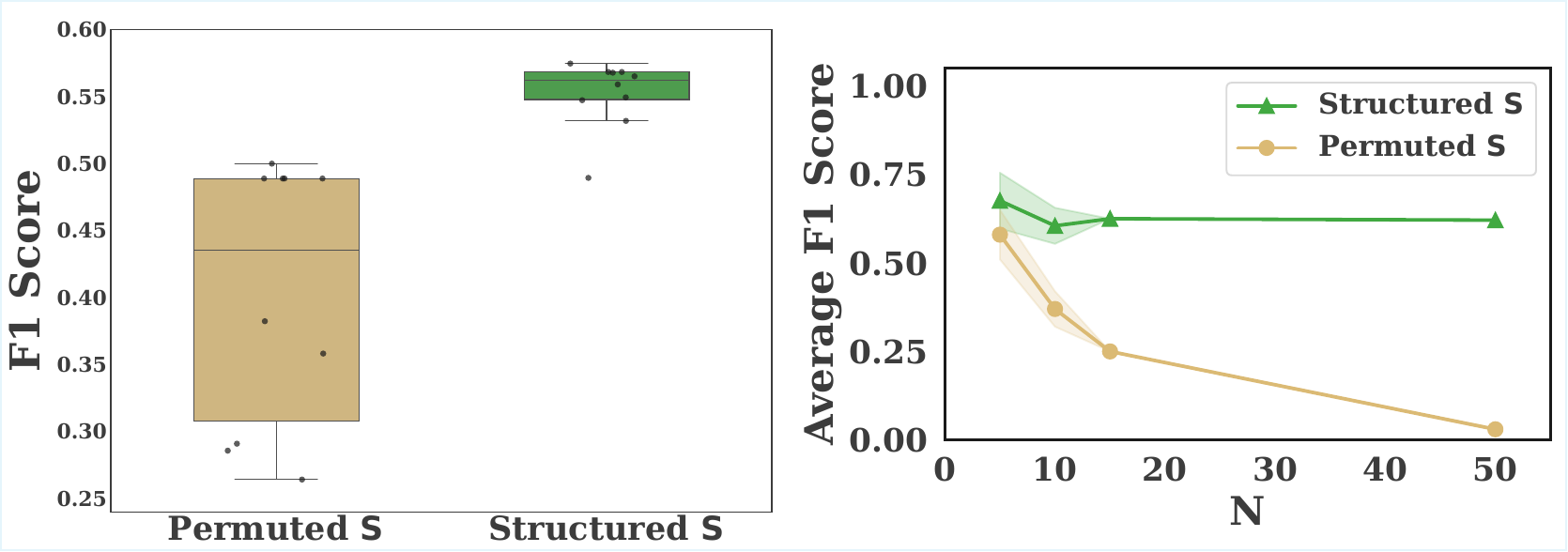}
  \caption{Impact of signal quality on F1 scores using original versus permuted causal signals on fMRI data.
  Performance degradation under signal corruption is minor for low-dimensional systems but substantial for higher-dimensional fMRI datasets.}
  \label{fig:randomS}
\end{figure}

Fig.~\ref{fig:randomS} shows that while Qbic can recover reasonable structure on low-dimensional systems (e.g., Lorenz-96 with fixed \( N=10 \)), performance degrades sharply on higher-dimensional fMRI datasets when signal structure is corrupted. This demonstrates that meaningful causal recovery depends critically on coherent intervention-induced signals, and cannot be attributed to sparsity selection alone.

\subsection{Effect of Clamping Magnitude}

We examine the sensitivity of Causal-INSIGHT to the choice of clamping magnitude \( x^* \in [0,1] \). Since all inputs are normalized, these values correspond to plausible input deviations. Across dataset categories, performance is stable over a broad range of values, with extremal clamping yielding the most consistent results. In particular, fMRI performance peaks near \( x^* = 0 \), aligning with a natural interpretation of biological inactivity, while synthetic systems benefit from stronger clamping near \( x^* = 1 \). Based on these observations, we adopt extremal clamping throughout, as it provides a simple and robust strategy across domains.

\section{Conclusion}

We present Causal-INSIGHT, a model-agnostic, post-hoc probing framework for extracting directed, time-lagged influence structure from trained temporal predictors. By analyzing a model’s response to controlled input clamping, the method recovers interpretable influence graphs that summarize the dependencies a predictor relies on for  prediction. Across synthetic, simulated, and realistic benchmarks, Causal-INSIGHT achieves competitive structural recovery while yielding statistically significant improvements in temporal delay localization, indicating that predictive temporal models implicitly encode stable directed dependencies that can be recovered post-hoc.

Overall, Causal-INSIGHT is best viewed as a complementary interpretation layer that enhances temporal precision while preserving the structural strengths of the underlying predictor, and can be applied uniformly across diverse architectures at inference time. However, the extracted graphs reflect model-implied dependencies rather than guarantees of the underlying data-generating causal structure. Without additional assumptions (e.g., correct model specification, causal sufficiency, and faithfulness), they should be interpreted as model-implied Granger-causal dependencies rather than identifiable structural causal models. Future work includes extending the probing framework to multi-horizon predictors and developing adaptive clamping schemes to better capture distributed or multi-lag influence patterns.

\section*{Acknowledgment}
The authors used ChatGPT (OpenAI) for limited language editing and stylistic suggestions \cite{chatgpt}. No scientific content, results, or conclusions were generated by the tool.

\normalsize
\bibliographystyle{IEEEtran}
\bibliography{refs}

\end{document}